\documentclass[runningheads]{llncs}
\usepackage[T1]{fontenc}
\usepackage{graphicx,verbatim}
\usepackage{amsmath}
\usepackage{multirow}
\usepackage{booktabs}
\usepackage{amssymb}
\usepackage{dsfont}
\usepackage{xcolor}
\usepackage{booktabs}
\usepackage{arydshln}
\usepackage{siunitx}
\begin{document}
\title{Learning from Sparse Point Labels for Dense Carcinosis Localization in Advanced Ovarian Cancer Assessment}
\titlerunning{Sparse Point Labels for Dense Carcinosis Localization}
%

\renewcommand{\thefootnote}{\fnsymbol{footnote}}
\footnotetext[1]{shared first/last authorship}
\def\thefootnote{$a$}\footnotetext{Corresponding author}
\def\thefootnote{$$}\footnotetext{\textit{This manuscript has been accepted for publication and will be included in the proceedings of MICCAI 2025.}}

\author{Farahdiba Zarin\inst{1}$^{a \star}$\and
Riccardo Oliva\inst{2,3,4}$^{\star}$\and
Vinkle Srivastav\inst{1,4} \and Armine Vardazaryan\inst{4} \and Andrea Rosati \inst{2} \and Alice Zampolini Faustini \inst{2}\and Giovanni Scambia \inst{2}\and Anna Fagotti \inst{2} \and Pietro Mascagni \inst{2,4}$^{\star}$\and Nicolas Padoy \inst{1,4}$^{\star}$}

%
\institute{University of Strasbourg, CNRS, INSERM, ICube, UMR7357, Strasbourg, France 
\email{fzarin@unistra.fr}
\and
Fondazione Policlinico Universitario A. Gemelli IRCCS, Università Cattolica del Sacro Cuore, Rome, Italy \and
IRCAD, Research Institute against Digestive Cancer, Strasbourg, France \and Institute of Image-Guided Surgery, IHU Strasbourg, Strasbourg, France \\ 
}
\authorrunning{F. Zarin et al.}


\maketitle              
\begin{abstract}
Learning from sparse labels is a challenge commonplace in the medical domain. This is due to numerous factors, such as annotation cost, and is especially true for newly introduced tasks. When dense pixel-level annotations are needed, this becomes even more unfeasible. However, being able to learn from just a few annotations at the pixel-level, while extremely difficult and underutilized, can drive progress in studies where perfect annotations are not immediately available. This work tackles the challenge of learning the dense prediction task of keypoint localization from a few point annotations in the context of 2d carcinosis keypoint localization from laparoscopic video frames for diagnostic planning of advanced ovarian cancer patients. To enable this, we formulate the problem as a sparse heatmap regression from a few point annotations per image and propose a new loss function, called Crag and Tail loss, for efficient learning. Our proposed loss function effectively leverages positive sparse labels while minimizing the impact of false negatives or missed annotations. Through an extensive ablation study, we demonstrate the effectiveness of our approach in achieving accurate dense localization of carcinosis keypoints, highlighting its potential to advance research in scenarios where dense annotations are challenging to obtain.


\end{abstract}
\section{Introduction}

Recent breakthroughs in surgical computer vision are driving the development of next-generation AI-assisted systems for the operating room~\cite{maier2022surgical}. The field has undergone significant evolution, transitioning from coarse-grained surgical workflow recognition~\cite{blum2008modeling} to more fine-grained surgical scene understanding tasks, including pixel-level scene segmentation~\cite{allan20192017}, surgical action localization~\cite{nwoye2022rendezvous}, and surgical scene reconstruction~\cite{rivoir2021long}. These advancements however have predominantly relied on fully supervised deep-learning models, which require extensive effort from clinical experts to generate accurately labeled ground truth data. Unlike natural images, which can be annotated by the general public, surgical annotations require specialized clinical knowledge. This is particularly challenging for dense pixel-level prediction tasks, where for example annotating a single image at the pixel level can take up to 90 minutes~\cite{cordts2016cityscapes}. Sparse annotations in comparison, are faster and easier to acquire. For example, in images containing multiple instances of similar entities, only a few examples can be annotated. Developing approaches that can effectively leverage sparse and noisy annotations for dense prediction tasks is therefore crucial, particularly in complex clinical applications where precise localization is essential.

One such application is the assessment of advanced ovarian cancer, one of the most lethal gynecologic malignancies due to its frequent late-stage diagnosis and extensive peritoneal dissemination~\cite{armstrong2021ovarian,siegel2018cancer,STEWART2019151}. Accurate intraoperative assessment during diagnostic laparoscopy is essential for determining cancer dissemination and guiding treatment decisions, which typically involve cytoreductive surgery (when deemed resectable) or platinum-based chemotherapy (when not)~\cite{fleming2021correlation,harrison2021cost}. The decision is taken assessing the Fagotti score or Peritoneal Cancer Index~\cite{fagotti2006laparoscopy,fagotti2011learning,fagotti2013multicentric}. The Fagotti score evaluates six anatomical stations: 1. Bowel, 2. Perietal Peritoneum, 3. Omentum, 4. Stomach, Lesser Omentum, and Spleen, 5. Diaphragm, 6. Liver and Gallbladder, for extent of carcinosis. Either $0$ or $2$ score is assigned to each station, 2 for extensive carcinosis spread~\cite{fagotti2006laparoscopy,fagotti2011learning,fagotti2013multicentric}. However, the current reliance surgeons assessment during diagnostic laparoscopy is time-consuming and subject to variability. AI-assisted analysis of laparoscopic videos could provide an objective and efficient way to quantify disease burden, facilitating the downstream clinical tasks of Fagotti score assessment, ultimately improving decision-making in treatment planning.

\begin{figure}[t!]
\centering
\includegraphics[width=\textwidth]{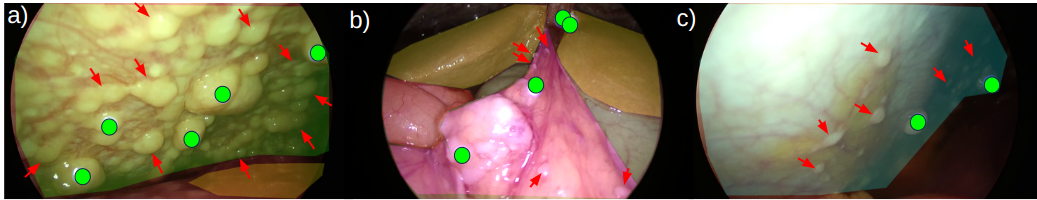}
\caption{Carcinosis appears as protruding nodules on the organ surfaces. Green points are sparse point labels from annotators. Red arrows points to more possible instances.} \label{fig1}
\end{figure}

In this work, we introduce a novel surgical computer vision task aimed at the dense localization of carcinosis keypoints in videos of advanced ovarian cancer captured during diagnostic laparoscopy. Given a few \emph{sparse} annotated keypoints per image during training, our goal is to perform \emph{dense} carcinosis localization in unseen diagnostic laparoscopic video (see Fig. 1). The task however is non-trivial and presents two challenges: few positive labels, and misleading incorrect labels. 

In general computer vision, point annotations are prevalent in human pose estimation, hand pose detection, and facial landmark detection tasks~\cite{cheng2020bottom,simon2017hand,sun2019deephighresolutionrepresentationlearning}. These typically use heatmap regression with Mean Squared Error (MSE) as the loss function~\cite{sun2019deephighresolutionrepresentationlearning,simon2017hand,cheng2020bottom,payer2016regressing,hervella2020deep}. While this works extremely well in a fully labeled setting, few tasks explore it with sparse labels. However, learning from sparse labels is well-explored for global prediction tasks such as classification and multilabel~\cite{song2022learning}. Out of existing strategies, loss reweighting~\cite{song2022learning,zhang2021simple} is a common and effective one  that is not limited by the requirements such as clean labels~\cite{mirikharaji2019learning,ren2018learning,Yi_2022,zhu2019pick} or geometric priors~\cite{johnson2011learning,liu2022explicit}. For instance, Hill loss specifically addresses missed annotations in multilabel tasks~\cite{zhang2021simple} by identifying false negatives and reducing their impact on the loss function. We hypothesize that adapting this loss function to dense prediction tasks could similarly improve learning from sparse point annotations. 

As our first contribution, we adapt the Hill loss function for dense prediction tasks by modifying it for heatmap regression. Through further analysis, we identify a limitation in the Hill loss function, specifically its tendency to down-weight the positive loss term. To address this, secondly, we propose a new loss function, termed \emph{Crag and Tail} loss, which introduces an additional term to reinforce the positive labels. Our proposed loss function achieves state-of-the-art performance. Finally, we conduct an extensive ablation study to assess the contribution of each component of the proposed loss function to the overall results. We describe our contributions as follows:\\
1. We adapt the Hill loss function for dense prediction tasks and identify its limitations, leading to the development of a novel \emph{Crag and Tail} loss function.\\
2. We achieve state-of-the-art performance and validate our approach through extensive ablation studies on the proposed loss function components.\\
3. We introduce a new surgical computer vision task for dense carcinosis localization in advanced ovarian cancer diagnostic laparoscopy videos using sparse keypoints.\\


\section{Methodology}

\begin{figure}[t!]
\centering
\includegraphics[width=\textwidth]{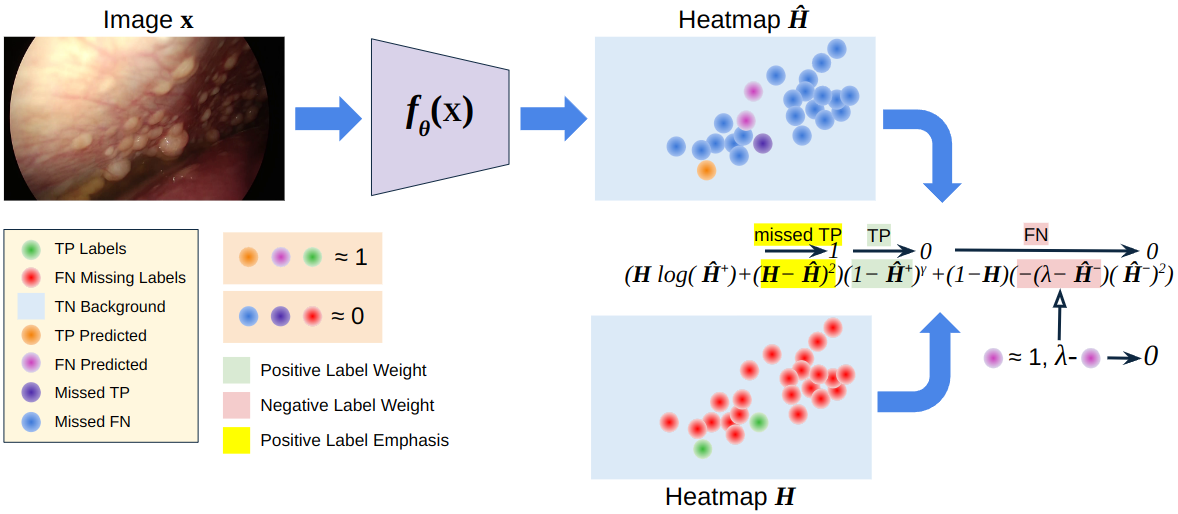}
\caption{Summarization of type of labels and loss components. False negatives (FN) with probabilities too similar to true positives (TP) have a low negative label weight, reducing their effect. The positive label term, since it focuses on hard-mining for positives, reduces the weight of easy positives. The reinforcement term doesn't remove the hard-mining for difficult TP samples, but rather adds a higher effect of TPs.} \label{fig2}
\end{figure}


\subsection{Problem Overview}

Consider a training dataset \(\mathcal{D} = \{ (\mathbf{X}, \mathbf{Y^*}) \}\), where \(\mathbf{X} \in \mathbb{R}^{3 \times h \times w}\) represents an RGB image with a spatial resolution of \(h \times w\), and \(\mathbf{Y^*} \in \mathbb{R}^{2 \times n}\) contains the 2D coordinates of \emph{sparse} carcinosis keypoints from \(n\) distinct locations. The goal is to train a deep learning model that can densely predict the 2D locations of all carcinosis keypoints across the entire image. Specifically, the model aims to learn a mapping function \(\mathcal{F}: \mathbf{X} \rightarrow \mathcal{\hat{H}} \in \mathbb{R}^{h \times w}\), which transforms an input image \(\mathbf{X}\) into a 2D heatmap \(\mathcal{\hat{H}}\). This heatmap encodes Gaussian distributions centered at each carcinosis keypoint, providing a probabilistic representation of their spatial locations.


\subsection{Learning from Sparse Point Annotations}
To train the model, we first generate a ground-truth heatmap \(\mathcal{H}\) from the sparse keypoints \(\mathbf{Y^*}\). Let \(\mathbf{x}_{j} \in \mathbb{R}^2\) denote the ground-truth position of the \(j^{th}\) carcinosis keypoint in \(\mathbf{Y^*}\). \(\mathcal{H}\), containing gaussian distributions centered at the sparsely annotated keypoints, is obtained by summing the individual heatmaps \(\mathcal{H}_{j}\) for all keypoints $\mathcal{H} = \sum_{j} \mathcal{H}_{j}$. The value at a location \(\mathbf{p} \in \mathbb{R}^2\) in the heatmap \(\mathcal{H}_{j}\) is defined as \(\mathcal{H}_{j}(\mathbf{p}) = \exp\left(-\frac{\|\mathbf{p} - \mathbf{x}_{j}\|_2^2}{\delta^2}\right)\), where \(\delta\) controls the spread of the gaussian peak.

Given the input image \(\mathbf{X}\) and the target heatmap \(\mathcal{H}\), we use an HRNet model~\cite{sun2019deephighresolutionrepresentationlearning} \(f_{\theta}(\cdot)\), parameterized by \(\theta\), to predict the output heatmap \(\mathcal{\hat{H}} = f_{\theta}(\mathbf{X})\). The standard MSE function to train the model can be defined as:
\( \mathcal{L}_{MSE} = \sum_{\mathbf{p}} \|\mathcal{\hat{H}}(\mathbf{p}) - \mathcal{H}(\mathbf{p})\|_2^2\). MSE is the most frequently used loss in heatmap regression tasks due to its robustness as a distance-based criterion for the pixel-level regression task over other losses~\cite{sun2019deephighresolutionrepresentationlearning,simon2017hand,cheng2020bottom,payer2016regressing,hervella2020deep}. However, the reason MSE works so well for heatmap regression in \emph{perfect} label settings, is also the limitation of this loss for \emph{sparsely} annotated keypoints. MSE heavily penalizes larger errors. This inadvertently means that false negatives, i.e. missing ground-truth points, are forced to match their incorrect ground truth. Therefore, large errors of false negative points, can overwhelm the learning, leading to suboptimal solutions. 

\textbf{Pixel-Level Uncertainty Aware Loss for Dense Prediction:}
To alleviate these pitfalls of MSE loss for missed labels, we propose a loss function that can tackle the false negatives. We start by reformulating the Hill Loss~\cite{zhang2021simple} for dense pixel-level regression tasks. Initially introduced for multi-label classification problems with missing labels, the adaptive weighting of Hill Loss places less emphasis on potential false negative samples, defined as follows:
\begin{equation}
\mathcal{L}_{Hill} = \sum_{\mathbf{p}} \mathcal{H} (1 - \mathcal{\hat{H}}^+)^\gamma \log(\mathcal{\hat{H}}^+) + (1 - \mathcal{H}) (-(\lambda -\mathcal{\hat{H}}^-)(\mathcal{\hat{H}}^{-})^2),
\end{equation}

where $\gamma$ is a focal parameter for weighting semi-hard positive samples, $\mathcal{\hat{H}}^+=\sigma(\mathcal{\hat{H}}-m$) and $\mathcal{\hat{H}}^-=\sigma(\mathcal{\hat{H}})$ are the heatmap components for the positive and negative loss terms, $m$ is a margin parameter for rescaling positive samples, $\lambda$ is a subtraction parameter for false negatives, and $\sigma(x) = \frac{1}{1+e^{-x}}$ is the sigmoid function. Here and afterwards, $\mathcal{H}(\mathbf{p})$, $\mathcal{\hat{H}}(\mathbf{p})$, $\mathcal{\hat{H}}^+(\mathbf{p})$, and $\mathcal{\hat{H}}^-(\mathbf{p})$ are referred to as $\mathcal{H}$, $\mathcal{\hat{H}}$, $\mathcal{\hat{H}}^+$, and $\mathcal{\hat{H}}^-$, respectively, for simplicity. 

The negative loss term \((1 - \mathcal{H}) (-(\lambda -\mathcal{\hat{H}}^-)(\mathcal{\hat{H}}^{-})^2)\) reduces the influence of potential false negatives by assigning smaller weights to samples with higher probabilities (closer to 1), which are likely to be false negatives during early training stages~\cite{zhang2021simple}. This allows the loss to intrinsically identify and downweight false negatives without explicit supervision. And, the positive loss term $\mathcal{H} (1 - \mathcal{\hat{H}}^+)^\gamma \log(\mathcal{\hat{H}}^+)$ emphasizes learning from semi-hard positives, ensuring they are prioritized during training~\cite{zhang2021simple}. 

\textbf{Reinforcement of Positive Labels:}
While the Hill loss effectively shifts focus toward easier samples with clear patterns, its down-weighting mechanism in the positive loss term, $\mathcal{H} (1 - \mathcal{\hat{H}}^+)^\gamma \log(\mathcal{\hat{H}}^+)$, inadvertently reduces emphasis on easier true positive samples. Additionally, the negative component of the loss, which is dominated by background pixels due to missing labels, results in low term values for false negatives (where $\mathcal{H} = 0$). Since true positives are valuable examples from which more can be learned, we introduce a new term, $(\mathcal{H} - \mathcal{\hat{H}})^2$, to increase attention on missed positive labels. Following these observations, we propose our \emph{Crag and Tail} loss, defined as:

\begin{equation}
\mathcal{L}_{CragAndTail} =  \sum_{\mathbf{p}} (\mathcal{H}\log(\mathcal{\hat{H}}^+)+(\mathcal{H} - \mathcal{\hat{H}})^2)(1 - \mathcal{\hat{H}}^+)^\gamma + (1 - \mathcal{H}) (-(\lambda -\mathcal{\hat{H}}^-)(\mathcal{\hat{H}}^{-})^2).
\end{equation}

By incorporating $(\mathcal{H} - \mathcal{\hat{H}})^2$, the proposed loss better captures and prioritizes true positives, effectively balancing the focus between true positives and potential false negatives, thus improving robustness to missing labels in the highly imbalanced setting typical of dense prediction tasks.

\textbf{Other Proposed Baseline Loss Functions:} We formulate different variations of the default MSE and Hill losses, namely $L_{SoftUncertainRegion}$, and $L_{0.5MaskedMSE}$, by explicit supervision of assumed false negatives. For \(n=0\), i.e. no carcinosis points in the image, we consider \(\mathcal{H}\) as all true negatives as no point annotations exist in the image. Similarly, for \(n>0\), true positives are contained within \(\mathcal{H}(\mathbf{p})>0\). Given that sparse points exist, it is reasonable to assume the existence of more unlabeled points. Outside of points, we assume the rest of the image of where \(\mathcal{H}(\mathbf{p})=0\) to be potential false negatives. These baseline losses are defined as follows:

\[
\mathcal{L}_{0.5MaskedMSE} =
\begin{cases} 
(\mathcal{H} - \mathcal{\hat{H}})^2, & \text{if } 
n=0 \text{ or } \mathcal{H}(\mathbf{p})>0  \\ 
a(\mathcal{H} - \mathcal{\hat{H}})^2, & \text{if }  n>0 \text{ and } \mathcal{H}(\mathbf{p})=0
\end{cases},
\]
\[
\mathcal{L}_{SoftUncertainRegionLoss} =
\begin{cases} 
\mathcal{L}_{MSE}+a\mathcal{L}_{Hill}, & \text{if } n=0 \text{ or } \mathcal{H}(\mathbf{p})>0 \\
a\mathcal{L}_{MSE}+\mathcal{L}_{Hill}, & \text{if } n>0 \text{ and } \mathcal{H}(\mathbf{p})=0
\end{cases}.
\]

These losses subdue the influence of these potential false negatives during training with a different down weighting scheme for each. We set $a=0.5$ for $\mathcal{L}_{SoftUncertainRegionLoss}$, and $\mathcal{L}_{0.5MaskedMSE}$.

\textbf{Heatmap Processing for Inference:} The predicted heatmaps are processed by performing non-maximum suppression
\(\mathcal{\hat{H}}(\mathbf{p})= \max_{(\mathbf{p}')\in \mathcal{N}(\mathbf{p})}\mathcal{\hat{H}}(\mathbf{p'})\).
Values of points that are not the highest valued in the local neighborhood $\mathcal{N}(\mathbf{p})$ of $5\times5$ are suppressed. The top-$k$ number of points are isolated using the threshold parameter $t$.

\section{Experimental Setup}

In this section, we briefly introduce the dataset, the evaluation metrics, and the implementation details of our experiments.

\textbf{Ovarian Laparoscopy Dataset:} 
The dataset consists of $30$ videos of diagnostic laparoscopy performed in patients with advanced stage ovarian cancer at Fondazione Policlinico Universitario Agostino Gemelli IRCCS (Rome, Italy), from 1st January to 31st December 2023. The local ethical committee approved the collection and analysis of these de-identified and pseudoanonymized data (IRB 6854/0022501/24). Corresponding clinical reports from the patients contained their diagnoses. Experienced surgeons from the gynecology department annotated keyframes for the Fagotti score assessment.

The six anatomical stations were annotated with segmentation masks. For a frame-level indication of whether the station contains carcinosis or not, 2D keypoint annotations are used for $1$-$5$ prominent cancerous keypoints inside the masks. The frames range from dimensions of $720$x$1280$ to $1080$x$1920$. Train, val, and test splits of 18:6:6 ratio were generated at video level  with the number of annotated frames, Fagotti Score, and station assessment as a reference to maintain uniform distribution. In total there are 1584 frames, 981 in train, 324 in val, and 279 in test. On average there are 5 points per annotated image.

\emph{Carcinosis Localization Evaluation:} Test frames were annotated for dense carcinosis with segmentation masks to locate every potential location that could contain points. 36 carcinosis instances exist on average per annotated test image.

\textbf{Evaluation:} For a comprehensive assessment of the losses and their contribution to the clinical task, we outline two evaluation schemes below:

\emph{Carcinosis Localization (Precision, Recall, F1):} We compute the true positives (TP) and false positives (FP) as the number of points inside or outside the carcinosis masks. The false negatives (FN) are indicated by the number of remaining individual carcinosis masks where no points are predicted.
The metrics per image are averaged. When the difference in precision is insignificant, the difference in recall is important as it highlights the capability to identify even more instances while maintaining the same level of correctness. Low precision means too many false positives even if recall is high.

\emph{Multilabel Classification (Precision, Recall, F1):} Given the assessment of presence or absence of carcinosis in an anatomical station, the task can be construed as multilabel classification. 
For each image, predicted heatmaps are processed to obtain the localized carcinosis coordinates. Any coordinates coinciding with a ground truth station mask indicate presence of carcinosis for that station. The binary presence or absence of carcinosis per station per image is used for the Precision, Recall, and F1, and the overall per station results averaged.


\textbf{Implementation Details:} All HRNet models are trained in batches of 4 with SGD of learning rate 0.01, L2 regularization of 0.0005, and a decay factor of 0.9 based on validation loss stagnation. For Hill loss and its adaptation, we use default parameters of $\lambda=1.5$, $\gamma=2$, and $m=1$. At inference, we use $k=30$, and $t=0.2$ for MSE loss, and $t=0.3$ for all other losses. Images are resized to $1280$x$720$. All models were trained on single A100s with early stopping based on validation loss.





\section{Results}

Here we outline the comparison of losses and weighing mechanisms, and ablation of the different loss components.

\textbf{Comparison of Losses:} As shown in Table~\ref{tab:results}, placing a smaller emphasis on potential false negatives in a naive fashion in $\mathcal{L}_{0.5MaskedMSE}$ greatly improves the normal MSE, proving the effect of false negatives to be detrimental. Our proposed $\mathcal{L}_{CragAndTail}$ achieves the highest recall and F1 scores for point localization, significantly outperforming $\mathcal{L}_{Hill}$, from which it is derived, across all metrics. When considering the trade-off between the metrics, our improvement in recall is considerably higher, indicating improved point localization.

\begin{table}[t!]
\caption{Precision, Recall, and F1 average of the different losses in point localization and multilabel classification. Tuned results of both $\mathcal{L}_{CragAndTail}$ and $\mathcal{L}_{Hill}$ with \((\lambda=1)\) are provided for fairness.}
\centering
\setlength{\tabcolsep}{6pt} 
\sisetup{table-format=2.2} 
\scalebox{0.8}{
\begin{tabular}{l *{3}{S} *{3}{S}} 
\toprule
\multirow{3}{*}{Loss} & \multicolumn{3}{c}{Point Localization} & \multicolumn{3}{c}{Multilabel Classification} \\  
\cmidrule(lr){2-4} \cmidrule(lr){5-7}
 & {Prec} & {Rec} & {F1} & {Prec} & {Rec} & {F1} \\  
\midrule
MSE & 54.72 & 15.83 & 18.99 & 74.77 & 36.27  & 48.22  \\
Hill & 73.90 & \underline{26.50} & \underline{35.23} & 78.29 & \underline{83.60}  & \underline{80.06} \\
Hill+MSE & 73.74 & 20.56 & 28.00 & \underline{83.04} & 76.97 & 79.24 \\
0.5 Masked MSE & 70.91 & 22.51 & 29.70 & 80.82  & 72.57  & 75.85 \\
Soft Uncertain Region & \textbf{76.68} & 21.55 & 28.52 & \textbf{83.76} & \ 74.19  & 78.11   \\
Crag and Tail & \underline{76.48} & \textbf{30.64} & \textbf{39.84} & 77.29 & \textbf{89.27}  & \textbf{81.84} \\
\midrule
Hill \((\lambda=1)\) & 73.89 & 44.01 & 52.06 & 69.71 & 91.09  & 78.03  \\
Crag and Tail \((\lambda=1)\) & 73.79 & 46.76 & 54.59 & 70.69 & 89.22  & 77.88  \\
\bottomrule
\end{tabular}
}
\label{tab:results}
\end{table}

Multilabel classification results measure the performance of point localization in the dedicated station masks. This highlights the contribution of our task in Fagotti score assessment, showing a considerable improvement in Recall and F1.

\textbf{Ablation of Loss Components:} As shown in Table~\ref{tab:loss_ablation}, analysis of the loss components highlights the importance of false positive suppression. Reducing the $\lambda$ to a too-small value significantly degrades the performance. Given the sparse nature of our annotations, where false negatives may have lower probabilities than in the multilabel task, reducing the $\lambda$ to $1$ is more effective than simply removing it. Reducing the semi-hard positive mining parameter $\gamma$ adversely affects performance. Similarly, removing the original positive Hill loss term and retaining only the reinforcement term also leads to a decline in performance.

\begin{table}[!t]
\caption{Ablation results of eliminating different loss components. Equations are placed following notation in Section 2.1.}
\centering
\setlength{\tabcolsep}{6pt} 
\sisetup{table-format=2.2} 
\scalebox{0.8}{ 
\begin{tabular}{p{4.5cm} *{3}{S}} 
\toprule
\multirow{2}{*}{Changes} & \multicolumn{3}{c}{Point Localization} \\  
\cmidrule(lr){2-4}
 & {Prec} & {Rec} & {F1}\\
\midrule
Default & \textbf{76.48} & 30.64 & 39.84 \\
$m=0$, no pos semi-hard mining & 75.24 & 25.26 & 33.69 \\
$m=0.5$ & 75.70 & 28.60 & 37.57 \\
$\gamma=0$, no weight in pos & 74.91 & 25.64 & 34.53\\
$\gamma=1$ & 74.69 & 28.57 & 37.59 \\
$\lambda=0$ & 9.91 & 28.27 & 13.05 \\
$\lambda=0.5$ & 12.04 & 32.95 & 15.90 \\
$\lambda=1$ & 73.79 & 46.76 & \textbf{54.59}  \\
$H\log(\hat{H}^+)=0$ & 16.13 & 11.91 & 11.98 \\
Only reinforce term in pos & 13.98 & 11.86 & 11.89\\
$(1 - \hat{H}^+)^\gamma=0$, only neg loss & 11.83 & 11.83 & 11.83 \\
$(\lambda -\hat{H^-})=1$, no weight in neg & 76.17 & 35.05 & 44.29 \\
$(\lambda -\hat{H^-})=0$, no neg Hill term & 31.64 & 62.23 & 36.74 \\
$H\log(\hat{H}^+)$, only pos Hill term & 33.60 & \textbf{63.05} & 38.27 \\
\bottomrule
\end{tabular}
}
\label{tab:loss_ablation}
\end{table}

\begin{figure}[t!]
\centering
\includegraphics[width=\textwidth]{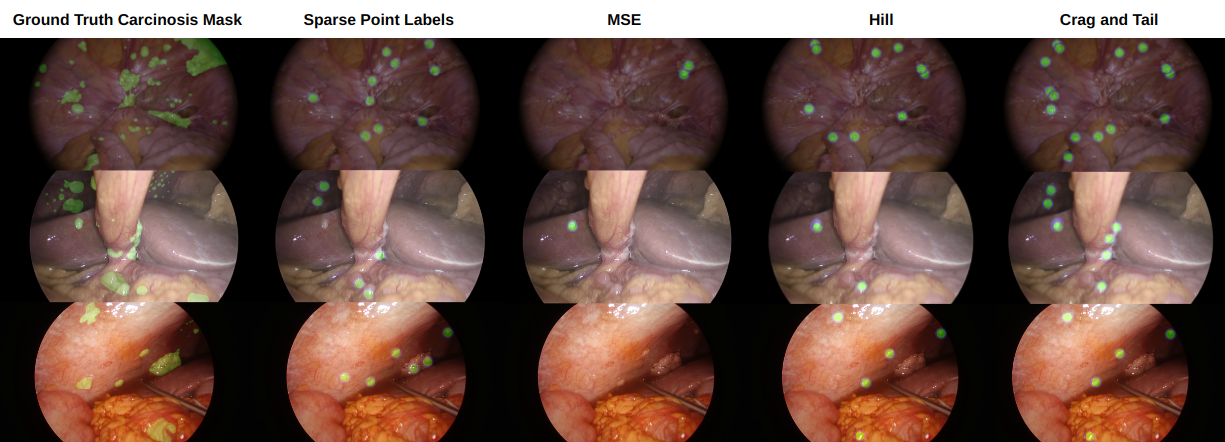}
\caption{Qualitative results of the loss performance compared. $\mathcal{L}_{CragAndTail}$ succeeds in localizing more carcinosis points. The traditional $\mathcal{L}_{MSE}$ struggles to learn from sparse point labels. For reference, the sparse point labels given by the clinicians are shown.} \label{qualitative_results}
\end{figure}

\section{Conclusion}

In this work, we overcome the difficulties of learning from a few correct pixel-level annotations for point localization. By thorough analysis of the Hill loss and adapting it for a novel clinical task, we propose our Crag and Tail loss by introducing a term in the positive loss designed to augment the effect of true positives further, while still preserving the benefits of Hill loss in suppressing false negatives. We highlight the robustness of Hill loss and our derived Crag and Tail loss thorough ablations of the losses for tackling a challenge prevalent in the medical domain.

    

\begin{credits}
\subsubsection{\ackname} This work has received funding from the European Union (ERC, CompSURG, 101088553). Views and opinions expressed are however those of the authors only and do not necessarily reflect those of the European Union or the European Research Council. Neither the European Union nor the granting authority can be held responsible for them. This work has also been supported by French state funds managed within the Plan Investissements d’Avenir by the ANR under reference ANR- 10-IAHU-02 (IHU Strasbourg). This work was granted access to the servers/HPC resources managed  by CAMMA, IHU Strasbourg, Unistra Mesocentre, and GENCI-IDRIS [Grant 2021- AD011011638R3, 2021-AD011011638R4].

\subsubsection{\discintname}
The authors have no competing interests to declare relevant to this article.
\end{credits}

%
%
%
\bibliographystyle{splncs04}
\bibliography{bibliography}
\end{document}